\documentclass{article}
\usepackage{iclr2026_conference,times}

\usepackage{amsmath,amsfonts,bm}

\def\eqref#1{equation~\ref{#1}}

\def\1{\bm{1}}

\def\eps{{\epsilon}}

\DeclareMathAlphabet{\mathsfit}{\encodingdefault}{\sfdefault}{m}{sl}
\SetMathAlphabet{\mathsfit}{bold}{\encodingdefault}{\sfdefault}{bx}{n}

\newcommand{\Ls}{\mathcal{L}}
\newcommand{\R}{\mathbb{R}}

\usepackage[colorlinks,citecolor=blue,linkcolor=blue,urlcolor=black]{hyperref}
\usepackage{xurl}
\usepackage{enotez}
\usepackage{graphicx}
\usepackage{svg}
\usepackage{subcaption}
\usepackage[capitalize,noabbrev]{cleveref}

\title{Optimal learning rate scaling depends on\\ data in deep scalar linear networks}

\author{Yedi Zhang\textsuperscript{1} \quad Peter E. Latham\textsuperscript{1} \quad Leena Chennuru Vankadara\textsuperscript{1} \quad Andrew Saxe\textsuperscript{1,2} \\
\textsuperscript{1}Gatsby Computational Neuroscience Unit, University College London\\
\textsuperscript{2}Sainsbury Wellcome Centre, University College London \\
\textsuperscript{ }\texttt{\{yedi,pel\}@gatsby.ucl.ac.uk,\{l.vankadara,a.saxe\}@ucl.ac.uk} \\
}

\iclrfinalcopy 
\begin{document}

\maketitle

\begin{abstract}
In this short note we consider the gradient descent dynamics of deep scalar linear networks, $f(x) = \prod_{l=1}^L w_l x$, which enjoy exact time-course solutions for any integer depth. We show that even in this minimal model, the optimal depth-wise learning rate scaling depends on data, whereas data-agnostic scaling rules fail to transfer across depths. Under the data-dependent optimal scaling, the learning dynamics is independent of data and weakly dependent on depth, resulting in a constant linear convergence rate across all depths including infinity. We further show similar data-dependent effects in deep scalar linear networks with residual connections.
\end{abstract}

\section{Introduction}
The large scale of modern neural networks has been empirically shown to play a crucial role in the rapid progress of deep learning models \citep{kaplan20scaling,hoffmann22chinchilla}. One essential factor of the scale is depth. Thus, understanding how to enable hyperparameter transfer across depth is critical for achieving predictable gains from scale. While existing literature on hyperparameter transfer suggests that data-agnostic learning rate scaling can allow depth-wise transfer \citep{yang24tensorvi,everett24exponent,noci24super,bordelon24depthwise,bordelon25transformer,bordelon25hyperparam}, we demonstrate that even in a minimal model class of deep scalar linear networks, the optimal learning rate scaling is inherently data-dependent. 
Specifically, the learning rate scaling rule contains a data-dependent correction that cannot be captured by a power law with a data-agnostic exponent, even when the constant factor is calibrated with data at one depth. We show that the learning rate scaling that accounts for this data-dependent exponent transfers across depth, whereas data-agnostic scaling does not.

We consider the simplest possible deep network, a depth-$L$ scalar linear chain defined as
\begin{align}  \label{eq:def-chain}
f(x; w) = \prod_{l=1}^L w_l x, \quad
x, w_1,\cdots,w_L \in \R.
\end{align}
Building on and refining analyses in prior work \citep{saxe13exact,saxe19semantic}, we write exact solutions to the full gradient descent learning dynamics for any integer depth, expressed via special functions, i.e. the hypergeometric function and the Lambert $W$ function. 
Under the data-dependent optimal learning rate scaling and a balanced initialization scheme, the gradient descent learning dynamics is independent of data and weakly dependent on depth. This results in a constant linear convergence rate across all depths, including the limiting case of infinite depth. 
Further, we extend the analysis to deep scalar linear residual networks with block depth one and two, and find that the optimal learning rate scaling for them is also data-dependent.

\textbf{Related work}.
\cite{jelassi23deeprelu} found that in deep ReLU networks with mean-field initialization, the largest learning rate for which the changes in the pre-activations after one gradient descent step remains bounded scales with depth $L$ as $L^{-3/2}$. \cite{bordelon24depthwise,bordelon25hyperparam} obtained reduced learning dynamics and studied hyperparameters transfer in infinite-depth linear residual networks in early training time, where the width and depth limits commute \citep{hayou23commute}. \cite{dey25dont} demonstrated that deep residual networks with $L^{-1/2}$ scaling can achieve hyperparameter transfer but operate in a locally lazy learning regime, while a $L^{-1}$ scaling enables rich learning and depth-wise hyperparameter transfer. 
Complementing these findings, we use a simple model class of deep scalar linear networks to demonstrate that the optimal learning rate scaling is data-dependent.

The learning dynamics of deep linear networks enjoy a rich line of theoretical results \citep{baldi89pca,fukumizu98batch,saxe13exact,saxe19semantic,arora18acc,shamir19converge,lampinen19gen,gidel19reg,advani20highd,huh20curvature,gissin20incremental,tarmoun21overparam,cengiz22silent,clem22prior,shi22pathways,yedi24unimodal,yedi25simplicity,clem25rich,xu25three,watanabe26anisotropic}. 
\cite{saxe13exact,saxe19semantic} solved the learning dynamics of deep linear networks with aligned small initial weights and white input covariance, showing that depth slows down learning in the case of learning with $\ell_2$ loss and the infinite-depth network incurs a finite decay in learning speed relative to the shallow network. Here we build on and refine these results by incorporating input correlations, expressing solutions via special functions, and choosing variables that reveal data-independent learning dynamics, and connect these results to the modern literature on hyperparameter transfer.

\section{Learning dynamics with the maximal stable learning rate }
Let $\{x_n, y_n\}_{n=1}^N$ be a training set. The gradient flow dynamics of the depth-$L$ linear chain in \cref{eq:def-chain} trained with $\ell_2$ loss, $\Ls=\frac1{2N} \sum_{n=1}^N(y-f(x))^2$, is given by
\begin{align}  \label{eq:gd}
\dot w_l = - \eta \frac{\partial \Ls}{\partial w_l}
= \eta \left(\mu_{yx} - \mu_{xx}  \prod_{i=1}^L w_i \right) \prod_{i\neq l} w_i ,
\end{align}
where $\mu_{yx} = \frac1N \sum_{n=1}^N y_n x_n, \mu_{xx}=\frac1N\sum_{n=1}^N x_n^2$ are moments of the dataset, and $\eta$ represents the learning rate. The continuous-time gradient flow dynamics captures the behaviors of gradient descent under a stable learning rate where the dynamics does not diverge or sustainedly oscillate \citep{cohen25centralflow}. We consider the stable regime of gradient descent learning in this paper. Here we present a self-contained exposition that builds on \cite{saxe13exact,saxe19semantic}, incorporating several refinements.

The dynamics in \cref{eq:gd} admits a well-known conservation law \citep{fukumizu98batch,saxe13exact,du18autobalance} between any pairs of weights, $\frac{d}{dt} \left( w_l^2 - w_{l'}^2 \right) = 0$.
If we assume all initial weights are positive\footnote{If the signs of the initial weights of each layer differ, the weights of some of the layers may change sign during learning. The $L$-dimensional dynamics is still constrained by the conservation law to evolve on a one-dimensional manifold, but we cannot write a one-dimensional differential equation to capture the full dynamics without using piecewise functions.}, 
$w_l(0)>0 \, \forall l$, we can use the conservation law to reduce the $L$-dimensional dynamics in \cref{eq:gd} to a one-dimensional ordinary differential equation about $w_1$
\begin{align}  \label{eq:1d-ode}
\dot w_1
= \eta \left(\mu_{yx} - \mu_{xx}  \prod_{i=1}^L \sqrt{w_1^2 + c_i} \right) \prod_{i=2}^L \sqrt{w_1^2 + c_i} ,\quad \text{where }
c_l = w_l(0)^2 - w_1(0)^2 .
\end{align}
We further assume that the initial weights of all layers are equal, $w_l(0) = w_1(0)\, \forall l$. This is motivated by the fact that we typically want all layers to participate in learning in a balanced way. Due to the conservation law, the weights that are initialized equal will remain equal throughout training. With the equal initial weight assumption, the dynamics in \cref{eq:1d-ode} simplifies to
\begin{align}
\dot w_1 = \eta \left(\mu_{yx} - \mu_{xx} w_1^L \right) w_1^{L-1} .
\end{align}

\begin{figure}
\includegraphics[width=\linewidth]{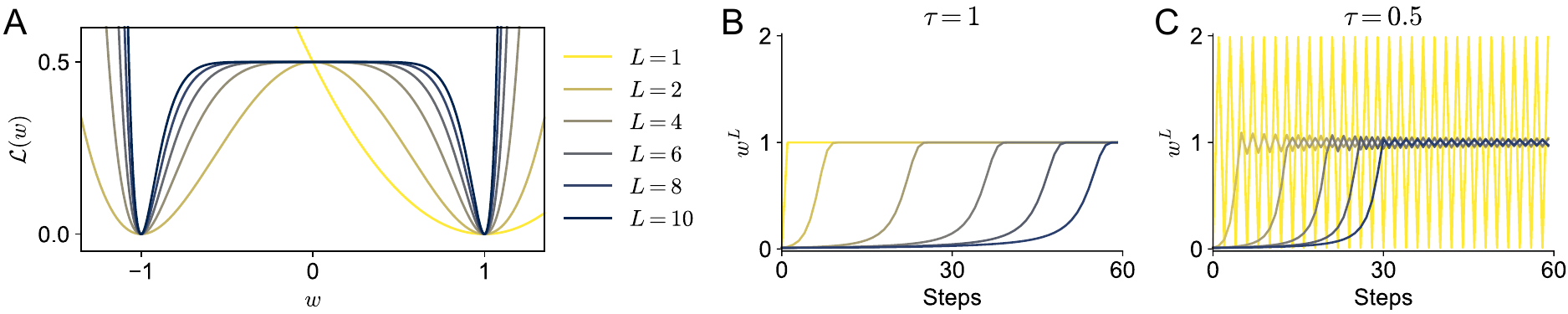}
\caption{The loss landscape of a scalar linear network has a sharper global minimum as the depth $L$ increases, requiring a smaller learning rate for stable gradient descent dynamics. (A) Plot of the loss function $\Ls(w)=\left(1-w^L \right)^2/2$ with different $L$. 
(B) Gradient descent trajectory of the total weight $w^L$ using learning rates that scale as \cref{eq:lr-scaling} with $\tau=1$. The dynamics is stable. (C) Same as panel B but with $\tau=0.5$, which is the threshold for stable gradient descent dynamics. The dynamics exhibits oscillations.}
\label{fig:landscape}
\end{figure}

\textbf{Maximum stable learning rate}.
When we increase the depth of the linear network, the global minimum of the loss landscape becomes sharper, as shown in \cref{fig:landscape}A. The second-order derivative, i.e. the sharpness, at the global minimum is $S=\mu_{xx} L \left( \frac{\mu_{yx}}{\mu_{xx}} \right)^{2-2/L}$.
A sharper minimum requires a smaller learning rate for gradient descent dynamics to be stable. In particular, the learning rate should satisfy: $0<\eta<2/S$. Hence, the maximum stable learning rate scales as
\begin{align}  \label{eq:lr-scaling}
\eta = \tau^{-1} \frac1{\mu_{xx} L} \left( \frac{\mu_{yx}}{\mu_{xx}} \right)^{-2+2/L} ,
\end{align}
where $\tau\in(0.5,\infty)$ is the time constant. The gradient descent dynamics exhibits oscillations when $\tau \leq 0.5$, as shown in \cref{fig:landscape}C. \cref{eq:lr-scaling} shows that even in this minimal setup, the scaling of the maximum stable learning rate is data-dependent, with implications on hyperparameter transfer that we will examine in \cref{sec:hp-transfer}.

\textbf{Dynamics with maximum stable learning rate}.
We now analyze the gradient flow dynamics with the maximum stable learning rate. We are interested in how the total weight evolves to approach the target weight over training. We thus study the dynamics of the ratio between the total weight and the target weight, $\alpha(t) = w_1(t)^L \mu_{xx}/\mu_{yx}$, which evolves as
\begin{align}  \label{eq:alpha-dyn}
\tau \dot \alpha = \alpha^{2-2/L} \left(1 - \alpha \right) . 
\end{align}
By using the maximum stable learning rate scaling and tracking the relative total weight rather than weights of individual layers, we obtain an ordinary differential equation (\ref{eq:alpha-dyn}) that is independent of the data statistics and weakly dependent on the depth $L$, i.e. through the factor $\alpha^{2-2/L}$. The dependence on $L$ weakens as $L$ increases, since $\lim_{L\to\infty}\alpha^{2-2/L}=\alpha^2$.

\textbf{Exact time-course solution}.
\cref{eq:alpha-dyn} is a separable differential equation. By separating variables and integrating both sides, we obtain the solution of $t$ in terms of $\alpha$ for any positive integer depth $L$
\begin{align}  \label{eq:sol-general}
t = \tau \frac{\alpha^{\frac2L-1}}{\frac2L-1} {}_2 F_1 \left(1,\frac2L-1;\frac2L;\alpha \right)\bigg|_{\alpha(0)}^{\alpha(t)} ,
\end{align}
where ${}_2 F_1$ is the hypergeometric function. For a general integer $L$, we cannot invert \cref{eq:sol-general} to solve $\alpha$ in terms of time $t$ due to the intractability of the hypergeometric function as a special function. However, we can invert \cref{eq:sol-general} for several specific depths, $L=1,2,\infty$, in which the hypergeometric function reduces to elementary functions. 
Specifically, in the limit of infinite-depth $L\to\infty$, the dynamics of $\alpha$ is given by 
\begin{align}  \label{eq:dyn-Linf}
\tau \dot \alpha = \alpha^2 \left(1 - \alpha \right) . 
\end{align}
The solution to \cref{eq:dyn-Linf} can be expressed as
\begin{align}  \label{eq:sol-Linf}
\alpha(t) = \frac1{1+W_0(e^{\beta(t)})} ,\quad \text{where } 
\beta(t) = - \frac{t}{\tau} + \frac1{\alpha_0} + \ln\left(\frac1{\alpha_0}-1 \right) - 1 ,\quad 0<\alpha\leq 1.
\end{align}
Here $W_0(\cdot)$ is the principal branch of the Lambert $W$ function. That is, $y=W_0(x)$ is the solution to the equation $y e^y=x$ with $x\geq0$. We provide the derivation for \cref{eq:sol-Linf} in \cref{supp:lambert}.

\textbf{Initial plateau}.
For any depth $L$, the dynamics in \cref{eq:alpha-dyn} has a stable fixed point at the global minimum, $\alpha=1$.
For deep networks, $L\geq2$, the network has an unstable fixed point at $\alpha=0$ in addition to the global minimum. If small initialization, the typical choice for feature learning \citep{woodworth20rich}, is used, the learning dynamics exhibits an initial plateau \citep{saxe13exact,saxe19semantic}, corresponding to slow escape from the zero fixed point.
The duration of the initial plateau $T$ is approximately
\begin{align}  \label{eq:plateau}
T = \tau \ln \frac1{\alpha(0)} \, , \text{ for } L=2;  \quad\quad
T = \frac{\tau}{(1-\frac2L) \alpha(0)^{1-\frac2L}} \, , \text{ for } L \geq 3.
\end{align}
The plateau duration $T$ increases when the depth $L$ increases and when the initialization $\alpha(0)$ decreases, as shown in \cref{fig:init-depth}. However, the plateau duration in the infinite-depth scalar linear network remains finite, with an upper bound of $T < \tau/\alpha(0)$.

\textbf{Convergence rate}.
At the end of learning, the linear scalar networks with $L\geq2$ all converge to the global minimum at a linear rate, according to the dynamics in \cref{eq:alpha-dyn}. That is, the total weight $\alpha$ is $\eps$-close to the global minimum, i.e. $|1-\alpha|<\eps$, after a time of order $\tau \ln \frac1\eps$.

\section{Depth-wise learning rate transfer  \label{sec:hp-transfer}}
For the scalar linear chain, the depthwise learning rate scaling rule in \cref{eq:lr-scaling} can be written as
\begin{align}
\eta_L = \tau^{-1} \frac{1}{\mu_{xx}L} r^{-2+2/L}, 
\quad \text{where }
r = \frac{\mu_{yx}}{\mu_{xx}} .
\end{align}
This is not a data-agnostic power law in \(L\): the factor \(L^{-1}\) is accompanied by a finite-depth correction that depends on the data statistics \(r\). To make this explicit, suppose the learning rate is tuned at a source depth \(L_0\). Transferring the learning rate from a source depth $L_0$ to a target depth $L$ requires
\begin{align}
\frac{\eta_L}{\eta_{L_0}}
= \frac{L_0}{L} r^{2(1/L-1/L_0)} .
\end{align}
The first factor, \(L_0/L\), is a data-agnostic scalar constant, while the second factor, $r^{2(1/L-1/L_0)},$ is data-dependent. Hence, a data-agnostic rule such as \(\eta_L/\eta_{L_0}=L_0/L\), even with a constant calibrated at the source depth, does not in general reproduce the finite-depth transfer rule. If one nevertheless fits the transfer rule between \(L_0\) and \(L\) with a
power law \(\eta_L/\eta_{L_0} = (L/L_0)^{-p_{\mathrm{eff}}}\), then
\begin{align}
p_{\mathrm{eff}}(L,L_0;r) =
1 - \frac{2(1/L-1/L_0)\ln r}{\ln(L/L_0)} .
\end{align}
Here the effective exponent $p_{\mathrm{eff}}$ depends on the data through \(r\). 

\begin{figure}[t]
\includegraphics[width=\linewidth]{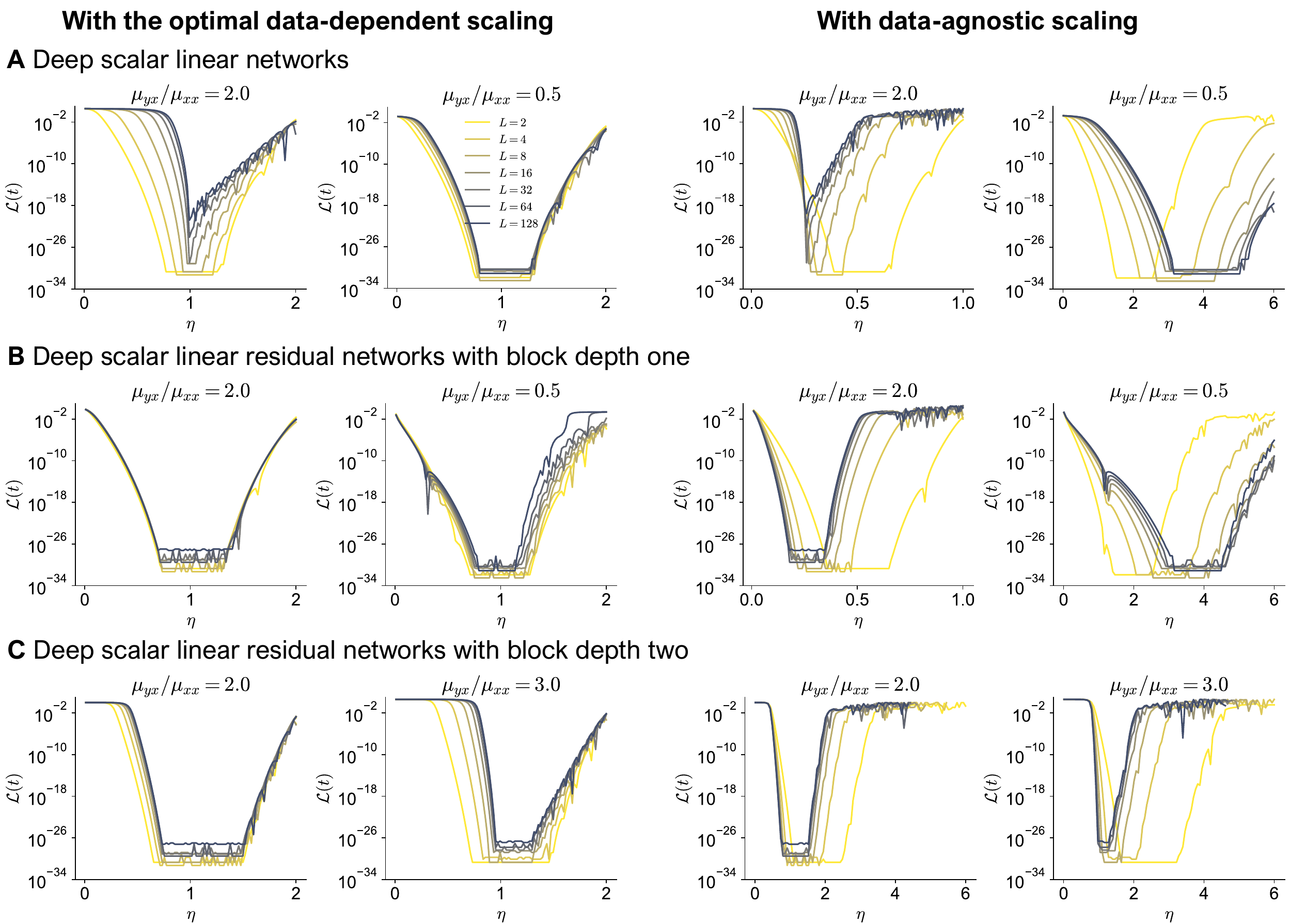}
\caption{Learning rates transfer under the optimal data-dependent scaling (left two columns), but not under data-agnostic scaling (right two columns). The optimal scaling for deep scalar linear networks, linear residual networks with block depth one and two are given by \cref{eq:lr-scaling,eq:resb1-lr-scaling,eq:resb2-lr-scaling}; the relevant data-agnostic scaling is $\eta\propto L^{-1}, 1, L$, respectively.
The loss values are the training loss after 30 steps of gradient descent. The initial weight is set to $w_l(0)^L=0.1$ in linear networks, and $w_l(0)=0.01$ in linear residual networks. In linear networks, we fix $w_l(0)^L$ rather than $w_l(0)$ across different depths because fixing $w_l(0)$ leads to vanishing gradients at initialization in the limit of large depth. In linear residual networks, a constant $w_l(0)$ suffices to avoid vanishing or exploding gradients at initialization.} 
\label{fig:hyperparam}
\end{figure}

Although our theory derives the maximal stable learning rate, the finite-horizon optimality is a related but different criterion. To address this gap, we empirically evaluate the finite-horizon optimality by sweeping learning rates and recording the training loss after a fixed number of gradient descent steps. In \cref{fig:hyperparam}A, we show the training loss after a fixed number of gradient descent steps  when the learning rate is scaled as the data-dependent rule in \cref{eq:lr-scaling} versus a data-agnostic power-law rule $\eta\propto L^{-1}$.
As shown in \cref{fig:hyperparam}A, the learning rates transfer from a shallower network to deep networks under the data-dependent scaling, but do not transfer under the data-agnostic scaling. Specifically, the learning rate with $L^{-1}$ scaling for a very deep network is too small when $\mu_{yx}/\mu_{xx}<1$, and too large when $\mu_{yx}/\mu_{xx}>1$.

We further extend the analysis to deep scalar linear residual networks with block depth one \citep{bordelon24depthwise,yang24tensorvi,marion25resnet} and block depth two \citep{bordelon25transformer,dey25dont}, defined as
\begin{align}
f_{\text{block1}}(x; w) = \prod_{l=1}^L \left(1+\frac{w_l}{\sqrt L} \right) x, 
\qquad \quad
f_{\text{block2}}(x; w) = \prod_{l=1}^L \left(1+\frac{w_l^2}L \right) x.
\end{align}
Their optimal learning rate scaling rules are calculated in \cref{eq:resb1-lr-scaling,eq:resb2-lr-scaling}, which are also data-dependent. Similar to deep scalar linear networks, the learning rates transfer under the optimal data-dependent scaling, but does not under data-agnostic scaling, as shown in \cref{fig:hyperparam}B,C.

On the flip side, we note that the data dependence of optimal learning rate scaling is weak for large $L$. Hence, transferring the optimal learning rate from an intermediate depth to large depth under $L^{-1}$ scaling may still suffice despite being suboptimal, whereas transferring the learning rate from a small depth (e.g. $L=2,4$) to large depth would likely fail, as we can see from \cref{fig:hyperparam}. 

In summary, we study depth-wise learning rate scaling in deep scalar linear networks, with and without residual connections, and find that the scaling rule for transfer is data-dependent.

\section*{Acknowledgments}
We thank Kevin Han Huang for feedback on a draft of this paper. We thank the following funding sources: Gatsby Charitable Foundation (GAT4058) to YZ, PEL, LCV, and AS; Sainsbury Wellcome Centre Core Grant from Wellcome (219627/Z/19/Z) to AS; Schmidt Science Polymath Award to AS. AS is a CIFAR Azrieli Global Scholar in the Learning in Machines \& Brains program.

\bibliography{ref}

@article{baldi89pca,
title = {Neural networks and principal component analysis: Learning from examples without local minima},
journal = {Neural Networks},
volume = {2},
number = {1},
pages = {53-58},
year = {1989},
issn = {0893-6080},
doi = {https://doi.org/10.1016/0893-6080(89)90014-2},
url = {https://www.sciencedirect.com/science/article/pii/0893608089900142},
author = {Pierre Baldi and Kurt Hornik}
}

@article{fukumizu98batch,
  title={Effect of batch learning in multilayer neural networks},
  author={Fukumizu, Kenji},
  journal={Gen},
  volume={1},
  number={04},
  pages={1E--03},
  year={1998}
}

@inproceedings{saxe13exact,
title={Exact solutions to the nonlinear dynamics of learning in deep linear neural networks},
author={Saxe, Andrew M and McClelland, James L and Ganguli, Surya},
booktitle={International Conference on Learning Representations},
year={2014},
url={https://arxiv.org/abs/1312.6120}
}

@inproceedings{poole16exponential,
 author = {Poole, Ben and Lahiri, Subhaneil and Raghu, Maithra and Sohl-Dickstein, Jascha and Ganguli, Surya},
 booktitle = {Advances in Neural Information Processing Systems},
 editor = {D. Lee and M. Sugiyama and U. Luxburg and I. Guyon and R. Garnett},
 pages = {},
 publisher = {Curran Associates, Inc.},
 title = {Exponential expressivity in deep neural networks through transient chaos},
 url = {https://proceedings.neurips.cc/paper_files/paper/2016/file/148510031349642de5ca0c544f31b2ef-Paper.pdf},
 volume = {29},
 year = {2016}
}

@inproceedings{schoenholz17infoprop,
title={Deep Information Propagation},
author={Samuel S. Schoenholz and Justin Gilmer and Surya Ganguli and Jascha Sohl-Dickstein},
booktitle={International Conference on Learning Representations},
year={2017},
url={https://openreview.net/forum?id=H1W1UN9gg}
}

@InProceedings{amos17optnet,
  title =    {{O}pt{N}et: Differentiable Optimization as a Layer in Neural Networks},
  author =       {Brandon Amos and J. Zico Kolter},
  booktitle =    {Proceedings of the 34th International Conference on Machine Learning},
  pages =    {136--145},
  year =   {2017},
  editor =   {Precup, Doina and Teh, Yee Whye},
  volume =   {70},
  series =   {Proceedings of Machine Learning Research},
  month =    {06--11 Aug},
  publisher =    {PMLR},
  url =    {https://proceedings.mlr.press/v70/amos17a.html}
}

@article{haber18stable,
doi = {10.1088/1361-6420/aa9a90},
url = {https://doi.org/10.1088/1361-6420/aa9a90},
year = {2017},
month = {dec},
publisher = {IOP Publishing},
volume = {34},
number = {1},
pages = {014004},
author = {Haber, Eldad and Ruthotto, Lars},
title = {Stable architectures for deep neural networks},
journal = {Inverse Problems}
}

@inproceedings{chen18neuralode,
 author = {Chen, Ricky T. Q. and Rubanova, Yulia and Bettencourt, Jesse and Duvenaud, David K},
 booktitle = {Advances in Neural Information Processing Systems},
 editor = {S. Bengio and H. Wallach and H. Larochelle and K. Grauman and N. Cesa-Bianchi and R. Garnett},
 pages = {},
 publisher = {Curran Associates, Inc.},
 title = {Neural Ordinary Differential Equations},
 url = {https://proceedings.neurips.cc/paper_files/paper/2018/file/69386f6bb1dfed68692a24c8686939b9-Paper.pdf},
 volume = {31},
 year = {2018}
}

@inproceedings{lampinen19gen,
title={An analytic theory of generalization dynamics and transfer learning in deep linear networks},
author={Andrew K. Lampinen and Surya Ganguli},
booktitle={International Conference on Learning Representations},
year={2019},
url={https://openreview.net/forum?id=ryfMLoCqtQ},
}

@InProceedings{xiao18isometry,
  title =    {Dynamical Isometry and a Mean Field Theory of {CNN}s: How to Train 10,000-Layer Vanilla Convolutional Neural Networks},
  author =       {Xiao, Lechao and Bahri, Yasaman and Sohl-Dickstein, Jascha and Schoenholz, Samuel and Pennington, Jeffrey},
  booktitle =    {Proceedings of the 35th International Conference on Machine Learning},
  pages =    {5393--5402},
  year =   {2018},
  editor =   {Dy, Jennifer and Krause, Andreas},
  volume =   {80},
  series =   {Proceedings of Machine Learning Research},
  month =    {10--15 Jul},
  publisher =    {PMLR},
  url =    {https://proceedings.mlr.press/v80/xiao18a.html}
}

@inproceedings{bai19deq,
 author = {Bai, Shaojie and Kolter, J. Zico and Koltun, Vladlen},
 booktitle = {Advances in Neural Information Processing Systems},
 editor = {H. Wallach and H. Larochelle and A. Beygelzimer and F. d\textquotesingle Alch\'{e}-Buc and E. Fox and R. Garnett},
 pages = {},
 publisher = {Curran Associates, Inc.},
 title = {Deep Equilibrium Models},
 url = {https://proceedings.neurips.cc/paper_files/paper/2019/file/01386bd6d8e091c2ab4c7c7de644d37b-Paper.pdf},
 volume = {32},
 year = {2019}
}

@Article{hanin19universal,
AUTHOR = {Hanin, Boris},
TITLE = {Universal Function Approximation by Deep Neural Nets with Bounded Width and ReLU Activations},
JOURNAL = {Mathematics},
VOLUME = {7},
YEAR = {2019},
NUMBER = {10},
ARTICLE-NUMBER = {992},
URL = {https://www.mdpi.com/2227-7390/7/10/992},
ISSN = {2227-7390},
DOI = {10.3390/math7100992}
}

@article{saxe19semantic,
author = {Andrew M. Saxe  and James L. McClelland  and Surya Ganguli },
title = {A mathematical theory of semantic development in deep neural networks},
journal = {Proceedings of the National Academy of Sciences},
volume = {116},
number = {23},
pages = {11537-11546},
year = {2019},
doi = {10.1073/pnas.1820226116},
URL = {https://www.pnas.org/doi/abs/10.1073/pnas.1820226116},
eprint = {https://www.pnas.org/doi/pdf/10.1073/pnas.1820226116}}

@InProceedings{shamir19converge,
  title =    {Exponential Convergence Time of Gradient Descent for One-Dimensional Deep Linear Neural Networks},
  author =       {Shamir, Ohad},
  booktitle =    {Proceedings of the Thirty-Second Conference on Learning Theory},
  pages =    {2691--2713},
  year =   {2019},
  editor =   {Beygelzimer, Alina and Hsu, Daniel},
  volume =   {99},
  series =   {Proceedings of Machine Learning Research},
  month =    {25--28 Jun},
  publisher =    {PMLR},
  url =    {https://proceedings.mlr.press/v99/shamir19a.html}
}

@inproceedings{gidel19reg,
 author = {Gidel, Gauthier and Bach, Francis and Lacoste-Julien, Simon},
 booktitle = {Advances in Neural Information Processing Systems},
 editor = {H. Wallach and H. Larochelle and A. Beygelzimer and F. d\textquotesingle Alch\'{e}-Buc and E. Fox and R. Garnett},
 pages = {},
 publisher = {Curran Associates, Inc.},
 title = {Implicit Regularization of Discrete Gradient Dynamics in Linear Neural Networks},
 url = {https://proceedings.neurips.cc/paper_files/paper/2019/file/f39ae9ff3a81f499230c4126e01f421b-Paper.pdf},
 volume = {32},
 year = {2019}
}

@inproceedings{gissin20incremental,
title={The Implicit Bias of Depth: How Incremental Learning Drives Generalization},
author={Daniel Gissin and Shai Shalev-Shwartz and Amit Daniely},
booktitle={International Conference on Learning Representations},
year={2020},
url={https://openreview.net/forum?id=H1lj0nNFwB}
}

@misc{kaplan20scaling,
      title={Scaling Laws for Neural Language Models}, 
      author={Jared Kaplan and Sam McCandlish and Tom Henighan and Tom B. Brown and Benjamin Chess and Rewon Child and Scott Gray and Alec Radford and Jeffrey Wu and Dario Amodei},
      year={2020},
      eprint={2001.08361},
      archivePrefix={arXiv},
      primaryClass={cs.LG},
      url={https://arxiv.org/abs/2001.08361}, 
}

@inproceedings{hoffmann22chinchilla,
 author = {Hoffmann, Jordan and Borgeaud, Sebastian and Mensch, Arthur and Buchatskaya, Elena and Cai, Trevor and Rutherford, Eliza and de Las Casas, Diego and Hendricks, Lisa Anne and Welbl, Johannes and Clark, Aidan and Hennigan, Thomas and Noland, Eric and Millican, Katherine and van den Driessche, George and Damoc, Bogdan and Guy, Aurelia and Osindero, Simon and Simonyan, Kar\'{e}n and Elsen, Erich and Vinyals, Oriol and Rae, Jack and Sifre, Laurent},
 booktitle = {Advances in Neural Information Processing Systems},
 editor = {S. Koyejo and S. Mohamed and A. Agarwal and D. Belgrave and K. Cho and A. Oh},
 pages = {30016--30030},
 publisher = {Curran Associates, Inc.},
 title = {An empirical analysis of compute-optimal large language model training},
 url = {https://proceedings.neurips.cc/paper_files/paper/2022/file/c1e2faff6f588870935f114ebe04a3e5-Paper-Conference.pdf},
 volume = {35},
 year = {2022}
}

@article{advani20highd,
title = {High-dimensional dynamics of generalization error in neural networks},
journal = {Neural Networks},
volume = {132},
pages = {428-446},
year = {2020},
issn = {0893-6080},
doi = {https://doi.org/10.1016/j.neunet.2020.08.022},
url = {https://www.sciencedirect.com/science/article/pii/S0893608020303117},
author = {Madhu S. Advani and Andrew M. Saxe and Haim Sompolinsky}}

@InProceedings{tarmoun21overparam,
  title = 	 {Understanding the Dynamics of Gradient Flow in Overparameterized Linear models},
  author =       {Tarmoun, Salma and Franca, Guilherme and Haeffele, Benjamin D and Vidal, Rene},
  booktitle = 	 {Proceedings of the 38th International Conference on Machine Learning},
  pages = 	 {10153--10161},
  year = 	 {2021},
  editor = 	 {Meila, Marina and Zhang, Tong},
  volume = 	 {139},
  series = 	 {Proceedings of Machine Learning Research},
  month = 	 {18--24 Jul},
  publisher =    {PMLR},
  url = 	 {https://proceedings.mlr.press/v139/tarmoun21a.html}
}

@inproceedings{clem22prior,
 author = {Braun, Lukas and Domin\'{e}, Cl\'{e}mentine and Fitzgerald, James and Saxe, Andrew},
 booktitle = {Advances in Neural Information Processing Systems},
 editor = {S. Koyejo and S. Mohamed and A. Agarwal and D. Belgrave and K. Cho and A. Oh},
 pages = {6615--6629},
 publisher = {Curran Associates, Inc.},
 title = {Exact learning dynamics of deep linear networks with prior knowledge},
 url = {https://proceedings.neurips.cc/paper_files/paper/2022/file/2b3bb2c95195130977a51b3bb251c40a-Paper-Conference.pdf},
 volume = {35},
 year = {2022}
}

@InProceedings{arora18acc,
  title = 	 {On the Optimization of Deep Networks: Implicit Acceleration by Overparameterization},
  author =       {Arora, Sanjeev and Cohen, Nadav and Hazan, Elad},
  booktitle = 	 {Proceedings of the 35th International Conference on Machine Learning},
  pages = 	 {244--253},
  year = 	 {2018},
  editor = 	 {Dy, Jennifer and Krause, Andreas},
  volume = 	 {80},
  series = 	 {Proceedings of Machine Learning Research},
  month = 	 {10--15 Jul},
  publisher =    {PMLR},
  url = 	 {https://proceedings.mlr.press/v80/arora18a.html}
}

@inproceedings{du18autobalance,
 author = {Du, Simon S and Hu, Wei and Lee, Jason D},
 booktitle = {Advances in Neural Information Processing Systems},
 editor = {S. Bengio and H. Wallach and H. Larochelle and K. Grauman and N. Cesa-Bianchi and R. Garnett},
 pages = {},
 publisher = {Curran Associates, Inc.},
 title = {Algorithmic Regularization in Learning Deep Homogeneous Models: Layers are Automatically Balanced},
 url = {https://proceedings.neurips.cc/paper_files/paper/2018/file/fe131d7f5a6b38b23cc967316c13dae2-Paper.pdf},
 volume = {31},
 year = {2018}
}

@inproceedings{shi22pathways,
 author = {Shi, Jianghong and Shea-Brown, Eric and Buice, Michael},
 booktitle = {Advances in Neural Information Processing Systems},
 editor = {S. Koyejo and S. Mohamed and A. Agarwal and D. Belgrave and K. Cho and A. Oh},
 pages = {34064--34076},
 publisher = {Curran Associates, Inc.},
 title = {Learning dynamics of deep linear networks with multiple pathways},
 url = {https://proceedings.neurips.cc/paper_files/paper/2022/file/dc3ca8bcd613e43ce540352b58d55d6d-Paper-Conference.pdf},
 volume = {35},
 year = {2022}
}

@inproceedings{cengiz22silent,
title={Neural Networks as Kernel Learners: The Silent Alignment Effect},
author={Alexander Atanasov and Blake Bordelon and Cengiz Pehlevan},
booktitle={International Conference on Learning Representations},
year={2022},
url={https://openreview.net/forum?id=1NvflqAdoom}
}

@InProceedings{woodworth20rich,
  title = 	 {Kernel and Rich Regimes in Overparametrized Models},
  author =       {Woodworth, Blake and Gunasekar, Suriya and Lee, Jason D. and Moroshko, Edward and Savarese, Pedro and Golan, Itay and Soudry, Daniel and Srebro, Nathan},
  booktitle = 	 {Proceedings of Thirty Third Conference on Learning Theory},
  pages = 	 {3635--3673},
  year = 	 {2020},
  editor = 	 {Abernethy, Jacob and Agarwal, Shivani},
  volume = 	 {125},
  series = 	 {Proceedings of Machine Learning Research},
  month = 	 {09--12 Jul},
  publisher =    {PMLR},
  url = 	 {https://proceedings.mlr.press/v125/woodworth20a.html}
}

@InProceedings{huh20curvature,
  title = 	 {Curvature-corrected learning dynamics in deep neural networks},
  author =       {Huh, Dongsung},
  booktitle = 	 {Proceedings of the 37th International Conference on Machine Learning},
  pages = 	 {4552--4560},
  year = 	 {2020},
  editor = 	 {III, Hal Daumé and Singh, Aarti},
  volume = 	 {119},
  series = 	 {Proceedings of Machine Learning Research},
  month = 	 {13--18 Jul},
  publisher =    {PMLR},
  url = 	 {https://proceedings.mlr.press/v119/huh20a.html}
}

@inproceedings{noci23shaped,
 author = {Noci, Lorenzo and Li, Chuning and Li, Mufan and He, Bobby and Hofmann, Thomas and Maddison, Chris J and Roy, Dan},
 booktitle = {Advances in Neural Information Processing Systems},
 editor = {A. Oh and T. Naumann and A. Globerson and K. Saenko and M. Hardt and S. Levine},
 pages = {54250--54281},
 publisher = {Curran Associates, Inc.},
 title = {The Shaped Transformer: Attention Models in the Infinite Depth-and-Width Limit},
 url = {https://proceedings.neurips.cc/paper_files/paper/2023/file/aa31dc84098add7dd2ffdd20646f2043-Paper-Conference.pdf},
 volume = {36},
 year = {2023}
}

@inproceedings{jacot23rank,
title={Implicit Bias of Large Depth Networks: a Notion of Rank for Nonlinear Functions},
author={Arthur Jacot},
booktitle={The Eleventh International Conference on Learning Representations },
year={2023},
url={https://openreview.net/forum?id=6iDHce-0B-a}
}

@InProceedings{hayou23commute,
  title = 	 {Width and Depth Limits Commute in Residual Networks},
  author =       {Hayou, Soufiane and Yang, Greg},
  booktitle = 	 {Proceedings of the 40th International Conference on Machine Learning},
  pages = 	 {12700--12723},
  year = 	 {2023},
  editor = 	 {Krause, Andreas and Brunskill, Emma and Cho, Kyunghyun and Engelhardt, Barbara and Sabato, Sivan and Scarlett, Jonathan},
  volume = 	 {202},
  series = 	 {Proceedings of Machine Learning Research},
  month = 	 {23--29 Jul},
  publisher =    {PMLR},
  url = 	 {https://proceedings.mlr.press/v202/hayou23a.html}
}

@article{hayou2023deep,
title={On the infinite-depth limit of finite-width neural networks},
author={Soufiane Hayou},
journal={Transactions on Machine Learning Research},
issn={2835-8856},
year={2023},
url={https://openreview.net/forum?id=RbLsYz1Az9},
note={}
}

@misc{jelassi23deeprelu,
      title={Depth Dependence of $\mu$P Learning Rates in ReLU MLPs}, 
      author={Samy Jelassi and Boris Hanin and Ziwei Ji and Sashank J. Reddi and Srinadh Bhojanapalli and Sanjiv Kumar},
      year={2023},
      eprint={2305.07810},
      archivePrefix={arXiv},
      primaryClass={cs.LG},
      url={https://arxiv.org/abs/2305.07810}, 
}

@InProceedings{yedi24unimodal,
  title = 	 {Understanding Unimodal Bias in Multimodal Deep Linear Networks},
  author =       {Zhang, Yedi and Latham, Peter E. and Saxe, Andrew M},
  booktitle = 	 {Proceedings of the 41st International Conference on Machine Learning},
  pages = 	 {59100--59125},
  year = 	 {2024},
  editor = 	 {Salakhutdinov, Ruslan and Kolter, Zico and Heller, Katherine and Weller, Adrian and Oliver, Nuria and Scarlett, Jonathan and Berkenkamp, Felix},
  volume = 	 {235},
  series = 	 {Proceedings of Machine Learning Research},
  month = 	 {21--27 Jul},
  publisher =    {PMLR},
  url = 	 {https://proceedings.mlr.press/v235/zhang24aa.html}
}

@inproceedings{yedi25simplicity,
title={Saddle-to-Saddle Dynamics Explains A Simplicity Bias Across Neural Network Architectures},
author={Yedi Zhang and Andrew M Saxe and Peter E. Latham},
booktitle={The Fourteenth International Conference on Learning Representations},
year={2026},
url={https://openreview.net/forum?id=Vit5M0G5Gb}
}

@article{marion25resnet,
  author  = {Pierre Marion and Adeline Fermanian and G{{\'e}}rard Biau and Jean-Philippe Vert},
  title   = {Scaling ResNets in the Large-depth Regime},
  journal = {Journal of Machine Learning Research},
  year    = {2025},
  volume  = {26},
  number  = {56},
  pages   = {1--48},
  url     = {http://jmlr.org/papers/v26/22-0664.html}
}

@inproceedings{bordelon25transformer,
 author = {Bordelon, Blake and Chaudhry, Hamza and Pehlevan, Cengiz},
 booktitle = {Advances in Neural Information Processing Systems},
 doi = {10.52202/079017-1130},
 editor = {A. Globerson and L. Mackey and D. Belgrave and A. Fan and U. Paquet and J. Tomczak and C. Zhang},
 pages = {35824--35878},
 publisher = {Curran Associates, Inc.},
 title = {Infinite Limits of Multi-head Transformer Dynamics},
 url = {https://proceedings.neurips.cc/paper_files/paper/2024/file/3eff068e195daace49955348de9f8398-Paper-Conference.pdf},
 volume = {37},
 year = {2024}
}

@inproceedings{bordelon24depthwise,
title={Depthwise Hyperparameter Transfer in Residual Networks: Dynamics and Scaling Limit},
author={Blake Bordelon and Lorenzo Noci and Mufan Bill Li and Boris Hanin and Cengiz Pehlevan},
booktitle={The Twelfth International Conference on Learning Representations},
year={2024},
url={https://openreview.net/forum?id=KZJehvRKGD}
}

@inproceedings{noci24super,
 author = {Noci, Lorenzo and Meterez, Alexandru and Hofmann, Thomas and Orvieto, Antonio},
 booktitle = {Advances in Neural Information Processing Systems},
 doi = {10.52202/079017-3262},
 editor = {A. Globerson and L. Mackey and D. Belgrave and A. Fan and U. Paquet and J. Tomczak and C. Zhang},
 pages = {102696--102743},
 publisher = {Curran Associates, Inc.},
 title = {Super Consistency of Neural Network Landscapes and Learning Rate Transfer},
 url = {https://proceedings.neurips.cc/paper_files/paper/2024/file/ba1d33849b963efc6b5d3082ad68f480-Paper-Conference.pdf},
 volume = {37},
 year = {2024}
}

@InProceedings{everett24exponent,
  title = 	 {Scaling Exponents Across Parameterizations and Optimizers},
  author =       {Everett, Katie E and Xiao, Lechao and Wortsman, Mitchell and Alemi, Alexander A and Novak, Roman and Liu, Peter J and Gur, Izzeddin and Sohl-Dickstein, Jascha and Kaelbling, Leslie Pack and Lee, Jaehoon and Pennington, Jeffrey},
  booktitle = 	 {Proceedings of the 41st International Conference on Machine Learning},
  pages = 	 {12666--12700},
  year = 	 {2024},
  editor = 	 {Salakhutdinov, Ruslan and Kolter, Zico and Heller, Katherine and Weller, Adrian and Oliver, Nuria and Scarlett, Jonathan and Berkenkamp, Felix},
  volume = 	 {235},
  series = 	 {Proceedings of Machine Learning Research},
  month = 	 {21--27 Jul},
  publisher =    {PMLR},
  url = 	 {https://proceedings.mlr.press/v235/everett24a.html}
}

@inproceedings{yang24tensorvi,
title={Tensor Programs {VI}: Feature Learning in Infinite Depth Neural Networks},
author={Greg Yang and Dingli Yu and Chen Zhu and Soufiane Hayou},
booktitle={The Twelfth International Conference on Learning Representations},
year={2024},
url={https://openreview.net/forum?id=17pVDnpwwl}
}

@inproceedings{cohen25centralflow,
title={Understanding Optimization in Deep Learning with Central Flows},
author={Jeremy Cohen and Alex Damian and Ameet Talwalkar and J Zico Kolter and Jason D. Lee},
booktitle={The Thirteenth International Conference on Learning Representations},
year={2025},
url={https://openreview.net/forum?id=sIE2rI3ZPs}
}

@inproceedings{clem25rich,
title={From Lazy to Rich: Exact Learning Dynamics in Deep Linear Networks},
author={Cl{\'e}mentine Carla Juliette Domin{\'e} and Nicolas Anguita and Alexandra Maria Proca and Lukas Braun and Daniel Kunin and Pedro A. M. Mediano and Andrew M Saxe},
booktitle={The Thirteenth International Conference on Learning Representations},
year={2025},
url={https://openreview.net/forum?id=ZXaocmXc6d}
}

@inproceedings{xu25three,
title={Three Mechanisms of Feature Learning in a Linear Network},
author={Yizhou Xu and Liu Ziyin},
booktitle={The Thirteenth International Conference on Learning Representations},
year={2025},
url={https://openreview.net/forum?id=Wh4SE2S7Mo}
}

@InProceedings{bordelon25hyperparam,
  title = 	 {Deep Linear Network Training Dynamics from Random Initialization: Data, Width, Depth, and Hyperparameter Transfer},
  author =       {Bordelon, Blake and Pehlevan, Cengiz},
  booktitle = 	 {Proceedings of the 42nd International Conference on Machine Learning},
  pages = 	 {4968--4997},
  year = 	 {2025},
  editor = 	 {Singh, Aarti and Fazel, Maryam and Hsu, Daniel and Lacoste-Julien, Simon and Berkenkamp, Felix and Maharaj, Tegan and Wagstaff, Kiri and Zhu, Jerry},
  volume = 	 {267},
  series = 	 {Proceedings of Machine Learning Research},
  month = 	 {13--19 Jul},
  publisher =    {PMLR},
  url = 	 {https://proceedings.mlr.press/v267/bordelon25a.html}
}

@inproceedings{dey25dont,
title={Don't be lazy: CompleteP enables compute-efficient deep transformers},
author={Nolan Simran Dey and Bin Claire Zhang and Lorenzo Noci and Mufan Li and Blake Bordelon and Shane Bergsma and Cengiz Pehlevan and Boris Hanin and Joel Hestness},
booktitle={The Thirty-ninth Annual Conference on Neural Information Processing Systems},
year={2025},
url={https://openreview.net/forum?id=lMU2kaMANl}
}

@misc{chizat25resnet,
      title={The Hidden Width of Deep ResNets: Tight Error Bounds and Phase Diagrams}, 
      author={Lénaïc Chizat},
      year={2025},
      eprint={2509.10167},
      archivePrefix={arXiv},
      primaryClass={cs.LG},
      url={https://arxiv.org/abs/2509.10167}, 
}

@misc{boix25resnet,
      title={On the inductive bias of infinite-depth ResNets and the bottleneck rank}, 
      author={Enric Boix-Adsera},
      year={2025},
      eprint={2501.19149},
      archivePrefix={arXiv},
      primaryClass={cs.LG},
      url={https://arxiv.org/abs/2501.19149}, 
}

@misc{watanabe26anisotropic,
      title={The Impact of Anisotropic Covariance Structure on the Training Dynamics and Generalization Error of Linear Networks}, 
      author={Taishi Watanabe and Ryo Karakida and Jun-nosuke Teramae},
      year={2026},
      eprint={2601.06961},
      archivePrefix={arXiv},
      primaryClass={stat.ML},
      url={https://arxiv.org/abs/2601.06961}, 
}
\bibliographystyle{iclr2026_conference}

\newpage
\appendix

\section{Additional related work}
A diverse body of theoretical research has investigated neural networks in the limit of large depth, regarding their expressivity \citep{poole16exponential,hanin19universal}, initialization scheme \citep{saxe13exact,schoenholz17infoprop,xiao18isometry,yang24tensorvi}, the network output at initialization \citep{hayou2023deep,noci23shaped}, the minimum-norm solution \citep{jacot23rank,boix25resnet}, and formulations based on implicit layers \citep{amos17optnet,bai19deq} and continuous-depth limits \citep{haber18stable,chen18neuralode}. 
Despite this progress, characterizing the behaviors of such deep networks once gradient descent training begins poses a greater challenge. 
Exact solutions for the full training dynamics have been derived for deep linear networks with aligned small initial weights and whitened data \citep{saxe13exact,saxe19semantic}.
For nonlinear networks, current findings characterize the gradient descent dynamics over only one or several steps \citep{jelassi23deeprelu,hayou2023deep,bordelon25hyperparam,bordelon25transformer,chizat25resnet}, while the full learning dynamics is generally intractable.

\section{Deep scalar linear networks}

\subsection{Additional figures}
\begin{figure}[h]
\centering
\includegraphics[width=0.24\linewidth]{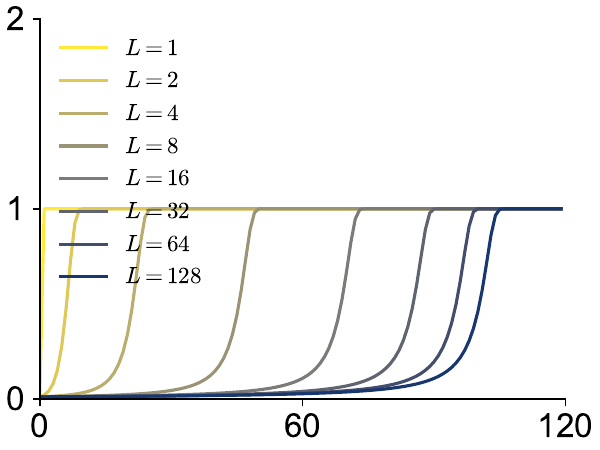}
\includegraphics[width=0.24\linewidth]{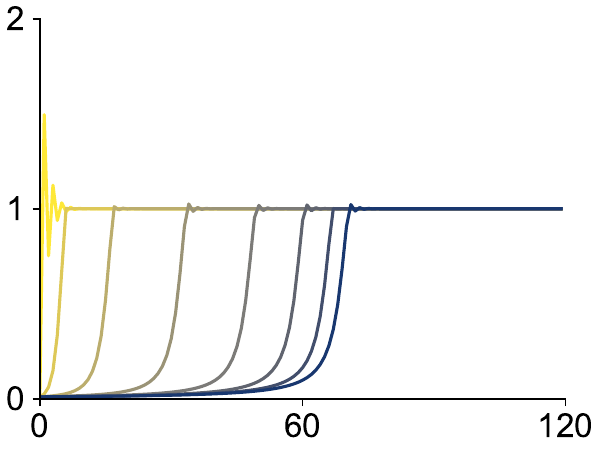}
\includegraphics[width=0.24\linewidth]{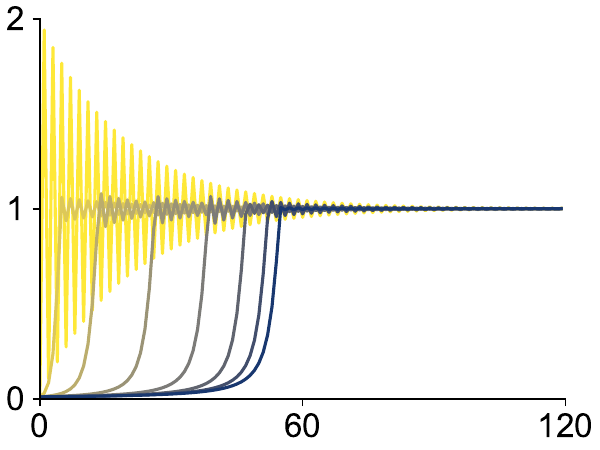}
\includegraphics[width=0.24\linewidth]{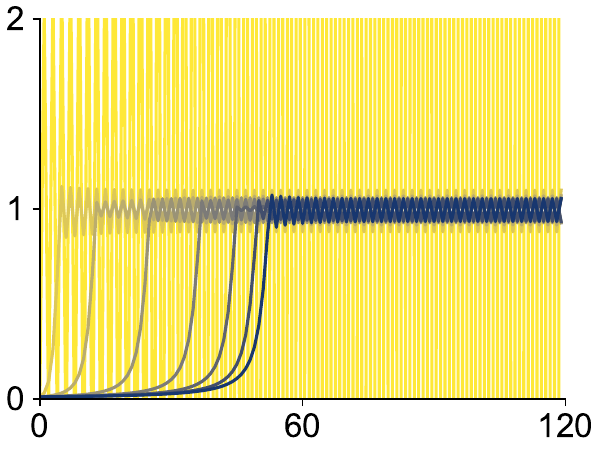}
\caption{Dynamics of $\alpha(t)$ with different depths $L$ and learning rates $\eta$. The learning rate $\eta$ is given by \cref{eq:lr-scaling} with $\tau^{-1} = 1,1.5,1.95,2.05$ for the four panels from left to right. When $0<\tau^{-1}\leq 1$, the gradient descent dynamics is monotonic and well described by the gradient flow dynamics. When $1<\tau^{-1}< 2$, the gradient descent dynamics is oscillatory but converging. When $\tau^{-1}\geq 2$, the gradient descent dynamics is oscillatory and diverging. Here the initialization is $\alpha(0)=0.01$. The data statistics are $\mu_{yx}=1, \mu_{xx}=1$.}
\label{fig:lr-depth}
\end{figure}

\begin{figure}[h]
\centering
\includegraphics[width=0.245\linewidth]{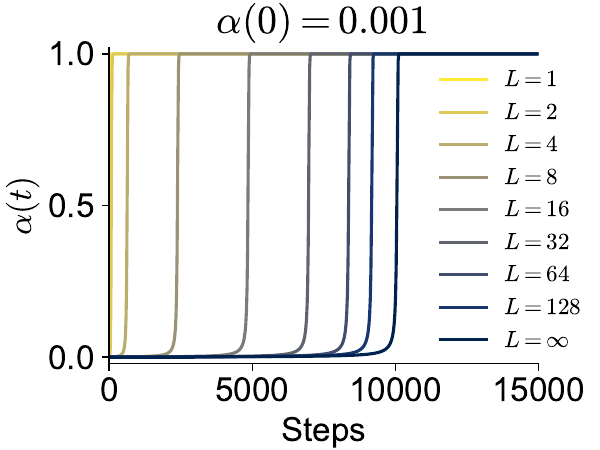}
\includegraphics[width=0.245\linewidth]{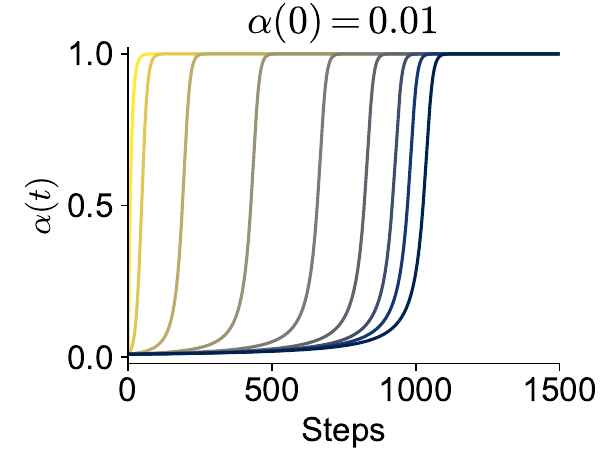}
\includegraphics[width=0.245\linewidth]{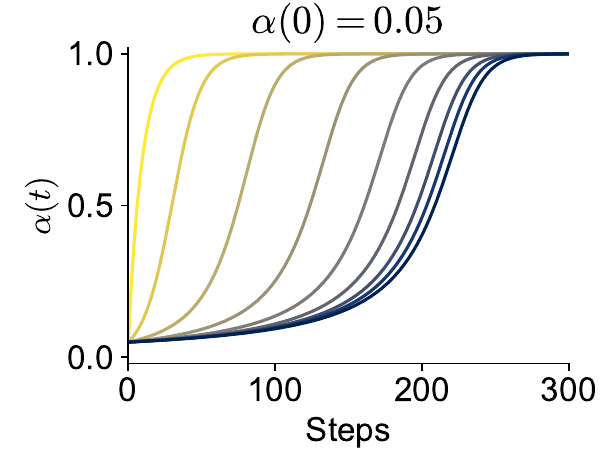}
\includegraphics[width=0.245\linewidth]{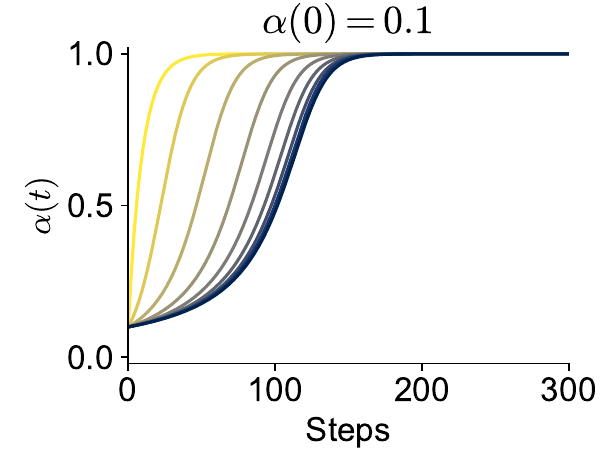}
\\ \vspace{1ex}
\includegraphics[width=0.245\linewidth]{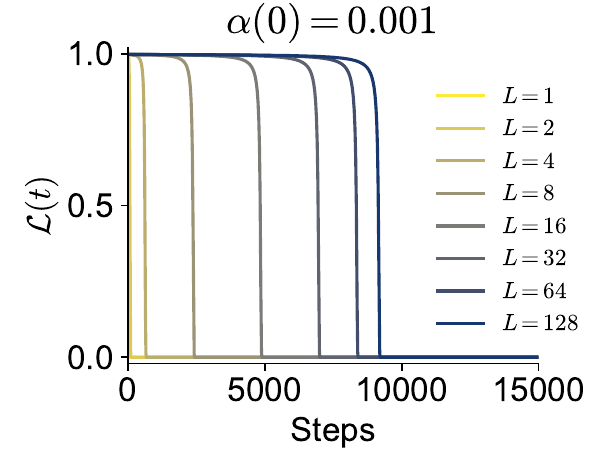}
\includegraphics[width=0.245\linewidth]{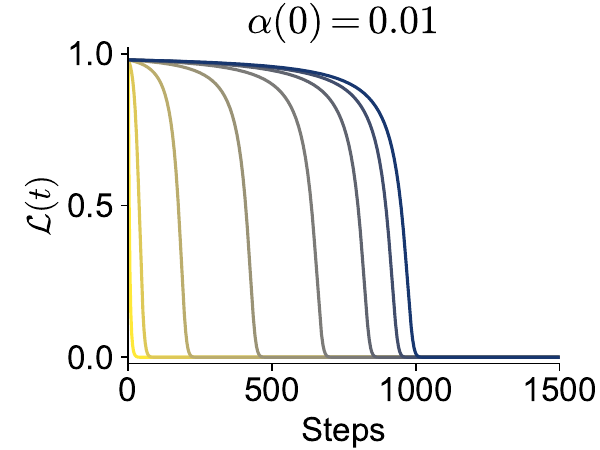}
\includegraphics[width=0.245\linewidth]{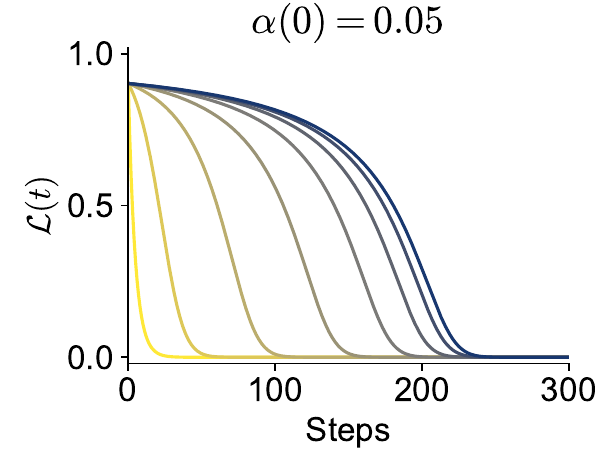}
\includegraphics[width=0.245\linewidth]{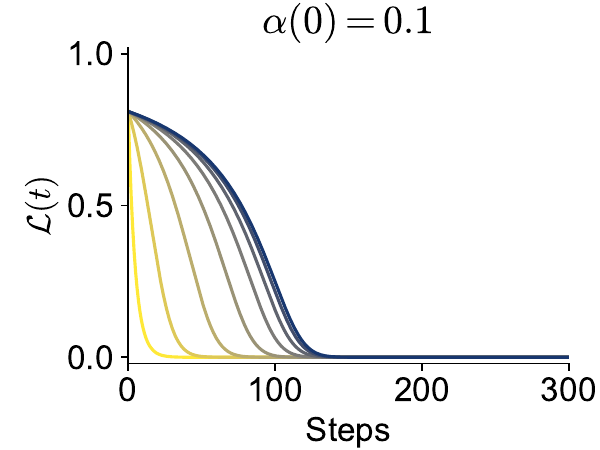}
\caption{Dynamics of total weights (top row) and loss (bottom row) with different depths $L$ and initialization $\alpha(0)$. The learning speed decreases when the depth increases and when the initialization scale decreases. Here the learning rate $\eta$ is given by \cref{eq:lr-scaling} with $\tau^{-1}=0.1$. The data statistics are $\mu_{yx}=1, \mu_{xx}=1$.}
\label{fig:init-depth}
\end{figure}

In \cref{fig:lr-depth}, we show the trajectories of the total weight with different depths and learning rates in deep scalar linear networks.
In \cref{fig:init-depth}, we show the trajectories of the total weight and loss with different depths and initialization in deep scalar linear networks.

\subsection{Derivation of the sharpness in \cref{eq:lr-scaling}}
By differentiating the gradient in \cref{eq:1d-ode} with respect to $w_1$, we obtain the sharpness of the loss landscape at which the weights in all layers are equal to $w_1$
\begin{align}
\frac1\eta \frac{\partial \dot w_1}{\partial w_1} 
= \mu_{yx} (L-1) w_1^{L-2} - \mu_{xx} (2L-1) w_1^{2L-2} .
\end{align}
The sharpness at the global minimum, denoted as $S$, is
\begin{align}  \label{eq:sharpness}
S \equiv \frac1\eta \frac{\partial \dot w_1}{\partial w_1} \bigg|_{w_1=\left(\frac{\mu_{yx}}{\mu_{xx}} \right)^{1/L}} 
= \mu_{xx} L \left( \frac{\mu_{yx}}{\mu_{xx}} \right)^{2-2/L} .
\end{align}
For $L=1$, the sharpness depends only on the input variance, $\mu_{xx}$, but not the input-output correlation, $\mu_{yx}$. For $L\geq2$, the sharpness depends on both the input variance and the input-output correlation.
We note that \cref{eq:sharpness} with $\mu_{xx}=1$ appeared in \citet[Equation (41)]{saxe13exact}.

Remark: 
When deriving the gradient descent dynamics, we calculate the negative gradient using the original loss expression with $L$ variables before substituting in the reduction $w_1=w_2=\cdots=w_L$. Substituting in the equality before taking the gradient would yield the wrong gradient descent dynamics.
However, when calculating the second-order derivative, we differentiate the expression in \cref{eq:1d-ode}, which is the gradient after substituting in the reduction $w_1=w_2=\cdots=w_L$. Substituting in the equality after the double differentiation would yield the wrong sharpness metric. This is because we want the sharpness of the loss landscape along the $w_1=w_2=\cdots=w_L$ path, not the sharpness along the $w_1$ axis with the rest of the weights held fixed.

\subsection{Derivation of of the total weight dynamics in \cref{eq:alpha-dyn}}
Using \cref{eq:1d-ode}, we obtain the dynamics of the total weight $a = w_1^L$
\begin{align}  \label{eq:a-dyn}
\dot a = \eta L a^{2-2/L} \left(\mu_{yx} - \mu_{xx} a \right) . 
\end{align}
\cref{eq:a-dyn} with $\mu_{xx}=1$ appeared in \citet[Equation (15)]{saxe13exact}.
Substituting the learning rate in \cref{eq:lr-scaling} into the dynamics of $a$, we get
\begin{align}
\tau \dot a = \left( \frac{\mu_{yx}}{\mu_{xx}} \right)^{-2+2/L} a^{2-2/L} \left(\frac{\mu_{yx}}{\mu_{xx}} - a \right) . 
\end{align}
We denote the total weight divided by the target weight as $\alpha(t) = w_1(t)^L \mu_{xx} / \mu_{yx}$, which represents the relative portion of the target weight learned, with $\alpha=1$ being the global minimum. The dynamics of $\alpha(t)$ is given by
\begin{align}
\tau \dot \alpha &= \tau \frac{\mu_{xx}}{\mu_{yx}} \dot a  \nonumber\\
&= \left( \frac{\mu_{xx}}{\mu_{yx}} a \right)^{2-2/L} \left(1 - \frac{\mu_{xx}}{\mu_{yx}} a \right)  \nonumber\\
&= \alpha^{2-2/L} (1-\alpha) .
\end{align}
We arrive at \cref{eq:alpha-dyn} in the main text.

\subsection{Derivation of the infinite-depth solution \cref{eq:sol-Linf}  \label{supp:lambert}}
We here solve the learning dynamics with $L\to\infty$, which is given by
\begin{align}
\tau \dot \alpha = \alpha^2 \left(1 - \alpha \right) . 
\end{align}
By separating variables and integrating both sides, we obtain
\begin{align}
\int_0^t \frac1\tau dt' &= \int_{\alpha_0}^{\alpha(t)} \frac{d\alpha'}{\alpha'^2(1-\alpha')}  \\
\Rightarrow \quad
\frac t\tau &= \left( - \frac1\alpha - \ln \left( \frac1\alpha-1\right)\right) \bigg|_{\alpha_0}^{\alpha(t)} .  \label{eq:inte-Linf}
\end{align}
\cref{eq:inte-Linf} appeared in \citet[Equation (17)]{saxe13exact}.
We rearrange \cref{eq:inte-Linf} and obtain
\begin{align}
\frac1{\alpha(t)} -1 + \ln \left( \frac1{\alpha(t)}-1\right) = - \frac{t}{\tau} + \frac1{\alpha_0} + \ln\left(\frac1{\alpha_0}-1 \right) - 1 \overset{\text{def}}= \beta(t) .
\end{align}
Taking the exponential of both sides yields
\begin{align}
\left( \frac1{\alpha(t)}-1\right) e^{\frac1{\alpha(t)} -1} = e^{\beta(t)} .
\end{align}
Because the principal branch of the Lambert $W$ function, denoted $y=W_0(x)$, solves the equation $y e^y=x$ with $x\geq0$, we have
\begin{align}
\frac1{\alpha(t)}-1 &= W_0(e^{\beta(t)}) \nonumber \\ 
\Rightarrow \quad
\alpha(t) &= \frac1{1+W_0(e^{\beta(t)})} , \quad 0 < \alpha \leq 1.
\end{align}
We arrive at \cref{eq:sol-Linf} in the main text.

\section{Deep scalar linear residual networks with block depth one  \label{supp:resb1}}
\begin{figure}[h]
\centering
\includegraphics[width=0.45\linewidth]{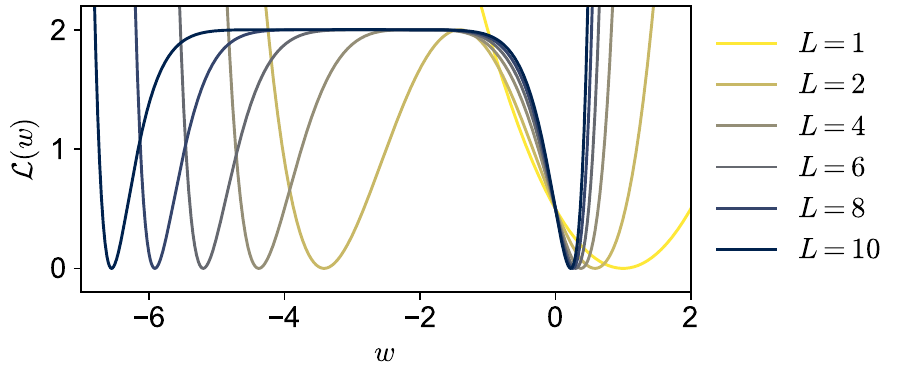}
\caption{The loss landscape of scalar linear residual networks with block depth one. Similar to the scalar linear chain in \cref{fig:landscape}, the sharpness of the global minimum increases with the depth $L$. Specifically, the plotted curves are $\Ls(w)=\left(2-(1+w/\sqrt L)^L \right)^2/2$, with different $L$. }
\label{fig:landscape-resb1}
\end{figure}

\begin{figure}[h]
\centering
\includegraphics[width=0.245\linewidth]{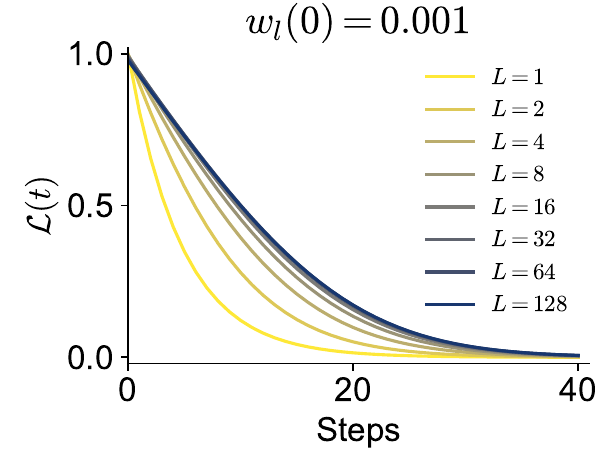}
\includegraphics[width=0.245\linewidth]{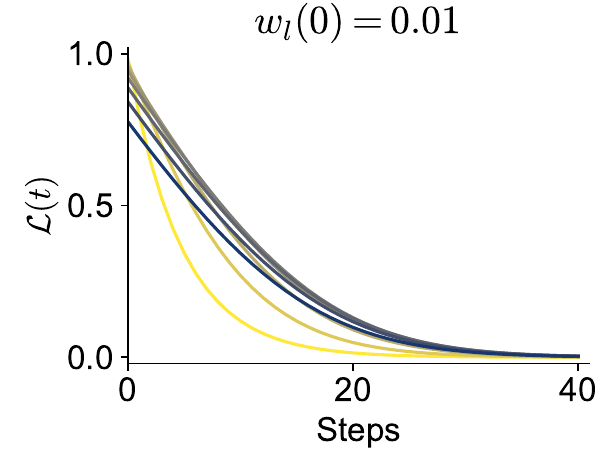}
\includegraphics[width=0.245\linewidth]{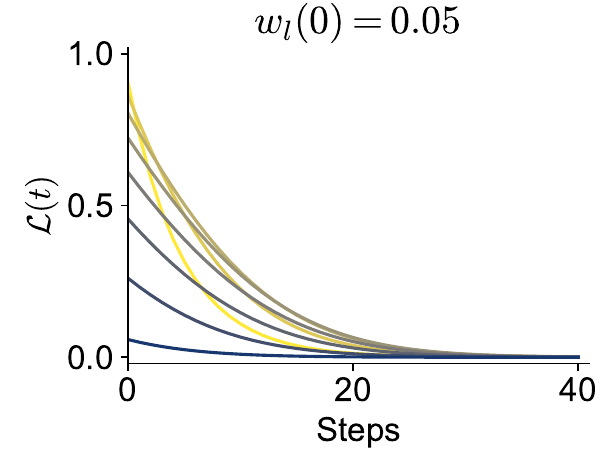}
\includegraphics[width=0.245\linewidth]{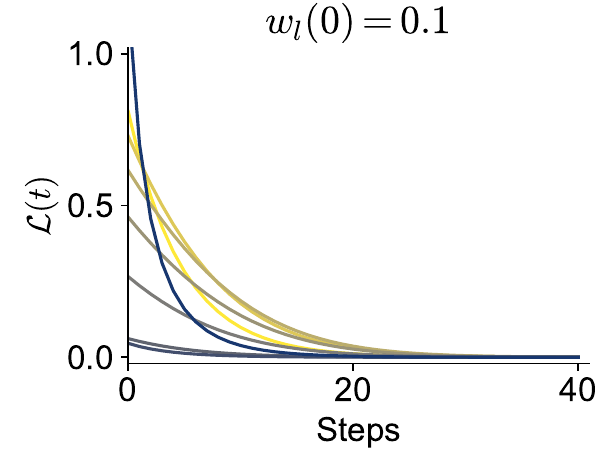}
\caption{Loss trajectories of deep scalar linear residual networks with block depth one with different depths and initialization. Here the learning rate $\eta$ is given by \cref{eq:resb1-lr-scaling} with $\tau^{-1}=0.1$. The data statistics are $\mu_{yx}=2, \mu_{xx}=1$.}
\end{figure}

Consider a scalar linear residual network with block depth one defined as 
\begin{align}
f(x; w) = \prod_{l=1}^L \left(1+\frac{w_l}{\sqrt L} \right) x, \quad
x, w_1,\cdots,w_L \in \R.
\end{align}
The $1/\sqrt L$ factor is a standard choice consistent with \cite{bordelon24depthwise,yang24tensorvi}.
The gradient flow dynamics trained with $\ell_2$ loss is given by
\begin{align}
\dot w_1 = \frac{\eta}{\sqrt L}\left[\mu_{yx} - \mu_{xx} \prod_{i=1}^L \left(1+\frac{w_i}{\sqrt L} \right) \right]
\prod_{i\neq l} \left(1+\frac{w_i}{\sqrt L} \right) .
\end{align}
Similar to the deep scalar linear network, we make the assumption of having equal initial weight in each layer, $w_l(0)=w_1(0)\, \forall l$, which will remain equal throughout training due to the conservation law. With equal weight in each layer, the gradient flow dynamics reduces to an one-dimensional ordinary differential equation
\begin{align}  \label{eq:resb1-1d-ode}
\dot w_1 = \frac{\eta}{\sqrt L}\left[\mu_{yx} - \mu_{xx} \left(1+\frac{w_1}{\sqrt L} \right)^L \right]
\left(1+\frac{w_1}{\sqrt L} \right)^{L-1} .
\end{align}
By differentiating the gradient in \cref{eq:resb1-1d-ode} with respect to $w_1$, we obtain the sharpness of the loss landscape at which the weights in all layers are equal to $w_1$
\begin{align}
\frac1\eta \frac{\partial \dot w_1}{\partial w_1}  
= \frac1{L} \left[ \mu_{yx} (L-1) \left(1+\frac{w_1}{\sqrt L} \right)^{L-2} - \mu_{xx} (2L-1) \left(1+\frac{w_1}{\sqrt L} \right)^{2L-2} \right] .
\end{align}
The sharpness at the global minimum is
\begin{align}
S \equiv \frac1\eta \frac{\partial \dot w_1}{\partial w_1}  \bigg|_{1+\frac{w_1}{\sqrt L} = \left(\frac{\mu_{yx}}{\mu_{xx}}\right)^{1/L}}
= \mu_{xx} \left( \frac{\mu_{yx}}{\mu_{xx}} \right)^{2-2/L} .
\end{align}
Hence, the maximum stable learning rate scales as
\begin{align}  \label{eq:resb1-lr-scaling}
\eta = \tau^{-1} \frac1{\mu_{xx}} \left( \frac{\mu_{yx}}{\mu_{xx}} \right)^{-2+2/L} ,
\end{align}
where $\tau\in(0.5,\infty)$ is the time constant.

As shown in \cref{fig:hyperparam}B, the optimal learning rate transfers under the data-dependent scaling in \cref{eq:resb1-lr-scaling}, but does not transfer under the data-agnostic constant scaling of $\eta\propto 1$.
Similar to deep scalar linear networks without residual connections, we note that the data dependence of the maximum stable learning rate is weak for large $L$ in deep scalar linear residual networks with block depth one, $\lim_{L\to\infty} 2/L=0$. Thus, transferring the optimal learning rate from an intermediate depth to infinite depth under the constant scaling is still justified, whereas transferring the learning rate from a small depth (e.g. $L=2,4$) to infinite depth would likely fail, as we can see from \cref{fig:hyperparam}B.

\section{Deep scalar linear residual networks with block depth two  \label{supp:resb2}}
\begin{figure}[h]
\centering
\includegraphics[width=0.45\linewidth]{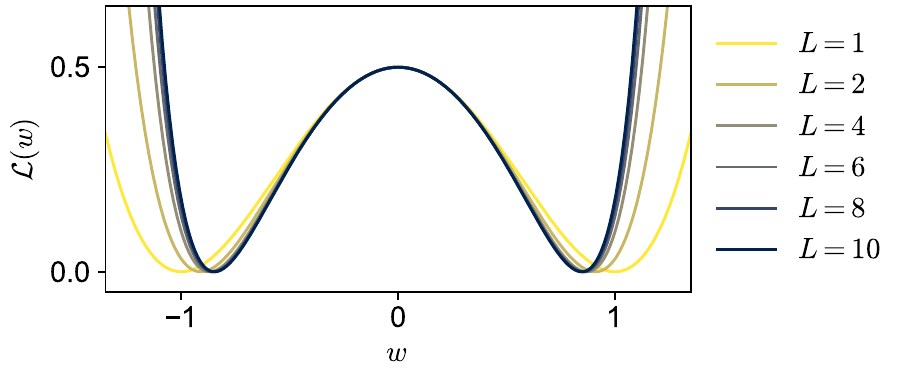}
\caption{The loss landscape of scalar linear residual networks with block depth two. Specifically, the plotted curves are $\Ls(w)=\left(2-(1+w^2/L)^L \right)^2/2$, with different $L$. }
\label{fig:landscape-resb2}
\end{figure}

\begin{figure}[h]
\centering
\includegraphics[width=0.245\linewidth]{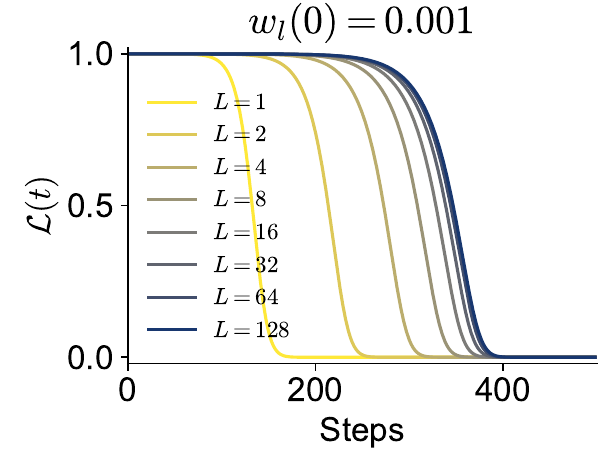}
\includegraphics[width=0.245\linewidth]{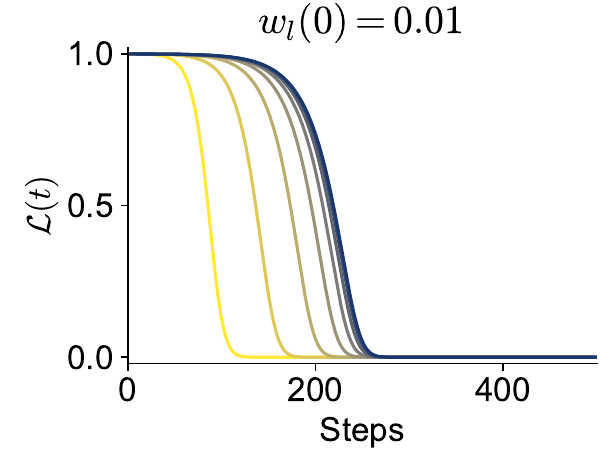}
\includegraphics[width=0.245\linewidth]{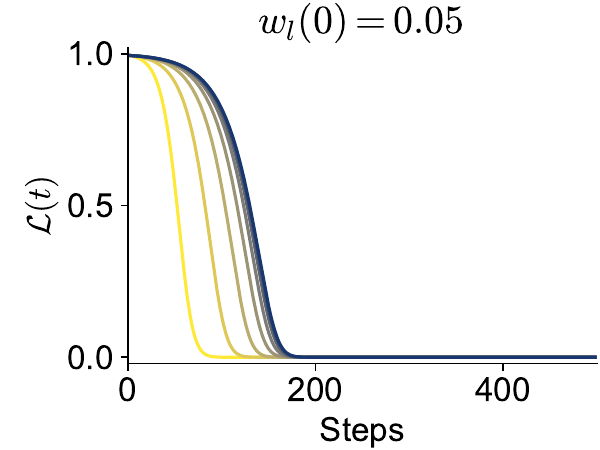}
\includegraphics[width=0.245\linewidth]{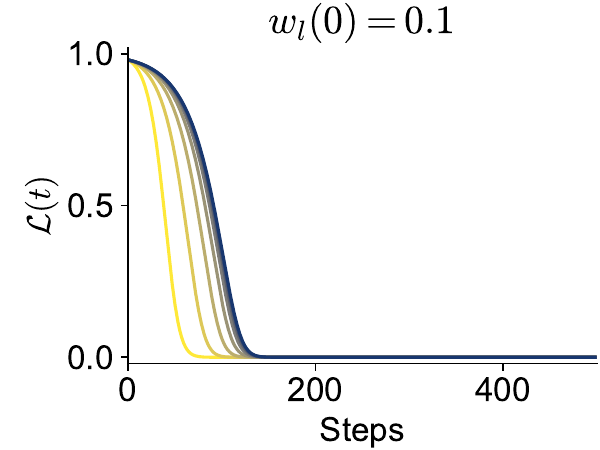}
\caption{Loss trajectories of deep scalar linear residual networks with block depth two with different depths and initialization. Here the learning rate $\eta$ is given by \cref{eq:resb2-lr-scaling} with $\tau^{-1}=0.1$. The data statistics are $\mu_{yx}=2, \mu_{xx}=1$.}
\end{figure}

Consider a scalar linear residual network with block depth two defined as
\begin{align}
f(x; w) = \prod_{l=1}^L \left(1+\frac{w_l^2}L \right) x, \quad
x, w_1,\cdots,w_L \in \R.
\end{align}
The $1/L$ factor is a standard choice consistent with \cite{bordelon25transformer,dey25dont}. In this architecture, we need $\mu_{yx}/\mu_{xx} \geq 1$, since the total weight $\prod_{l=1}^L \left(1+\frac{w_l^2}L \right) \geq 1$.
The gradient flow dynamics trained with $\ell_2$ loss is given by
\begin{align}
\dot w_1 = \frac{2\eta w_1}{L} \left[\mu_{yx} - \mu_{xx} \prod_{i=1}^L \left(1+\frac{w_i^2}L \right) \right]
\prod_{i\neq l} \left(1+\frac{w_i^2}L \right) .
\end{align}
Similar to the deep scalar linear network, we make the assumption of having equal initial weight in each layer, $w_l(0)=w_1(0)\, \forall l$, which will remain equal throughout training due to the conservation law. With equal weight in each layer, the gradient flow dynamics reduces to an one-dimensional ordinary differential equation
\begin{align}
\dot w_1 = \frac{2\eta}L \left[\mu_{yx} - \mu_{xx} \left(1+\frac{w_1^2}L \right)^L \right]
\left(1+\frac{w_1^2}L \right)^{L-1} w_1 .
\end{align}
By differentiating the gradient in \cref{eq:resb1-1d-ode} with respect to $w_1$, we obtain the sharpness of the loss landscape at which the weights in all layers are equal to $w_1$
\begin{align}
\frac1\eta \frac{\partial \dot w_1}{\partial w_1} 
= \frac{2}{L} \left[
  \mu_{yx} \left(1+\frac{w_1^2}L \right)^{L-2} \left(\frac{2L-1}L w^2 + 1 \right) 
- \mu_{xx} \left(1+\frac{w_1^2}L \right)^{2L-2} \left(\frac{4L-1}L w^2 + 1 \right)
\right] .
\end{align}
The sharpness at the global minimum is 
\begin{align}  \label{eq:resb2-sharpness}
S \equiv \frac1\eta 
\frac{\partial \dot w_1}{\partial w_1}  \bigg|_{1+\frac{w_1^2}L = \left(\frac{\mu_{yx}}{\mu_{xx}}\right)^{1/L}}
&= \frac{4}{L} \mu_{xx} \left(\frac{\mu_{yx}}{\mu_{xx}} \right)^{2-2/L} w^2  \nonumber\\
&= 4\mu_{xx} \left(\frac{\mu_{yx}}{\mu_{xx}} \right)^{2-2/L} \left(\left(\frac{\mu_{yx}}{\mu_{xx}}\right)^{1/L}-1 \right) 
\end{align}
Hence, the maximum stable learning rate scales as
\begin{align}  \label{eq:resb2-lr-scaling}
\eta = \tau^{-1} \frac1{4\mu_{xx}} \left(\frac{\mu_{yx}}{\mu_{xx}} \right)^{-2+2/L} \left(\left(\frac{\mu_{yx}}{\mu_{xx}}\right)^{1/L}-1 \right)^{-1}
\end{align}
where $\tau\in(0.5,\infty)$ is the time constant.

The scaling of \cref{eq:resb2-sharpness} with respect to $L$ is not immediately apparent. To see its behavior with large $L$, we Taylor expand \cref{eq:resb2-sharpness} around $1/L=0$, which yields
\begin{align}
S = \frac4L \mu_{xx} \left(\frac{\mu_{yx}}{\mu_{xx}} \right)^2 \ln \left(\frac{\mu_{yx}}{\mu_{xx}} \right) + O \left(\frac1{L^2}\right) .
\end{align}
This shows that the sharpness at the global minimum decreases with $L$, scaling as $1/L$. Therefore, if we were to use a data-agnostic power-law scaling, the learning rate would scale with depth as $\eta\propto L$.

In \cref{fig:hyperparam}C, we compare the learning rate transfer between the exact maximum stable learning rate scaling in \cref{eq:resb2-lr-scaling} and the data-agnostic scaling of $\eta\propto L$. Similar to the cases with deep scalar linear networks and scalar linear residual networks with block depth one, the optimal learning rate transfers under the data-dependent scaling in \cref{eq:resb2-lr-scaling}, but not under the data-agnostic scaling of $\eta\propto L$.

\section*{Reproducibility}
Code reproducing our figures is available at GitHub: \href{https://github.com/yedizhang/scalar-chain}{\texttt{github.com/yedizhang/scalar-chain}}

\end{document}